%% file: main.tex
\pdfoutput=1

\documentclass[11pt]{article}

\usepackage[preprint]{acl}

\usepackage{times}
\usepackage{latexsym}

\usepackage[T1]{fontenc}

\usepackage[utf8]{inputenc}

\usepackage{microtype}

\usepackage{inconsolata}

\usepackage{graphicx}

\usepackage{amsmath}
\usepackage{tabularx}
\usepackage{booktabs}

\usepackage{colortbl} 
\usepackage{amssymb}
\usepackage{multirow}
\usepackage{fontawesome}

\newcommand{\ours}{ACD}
\newcommand{\cu}{MICD}

\newcommand{\astfootnote}[1]{
    \let\oldthefootnote=\thefootnote
    \setcounter{footnote}{1}
    \renewcommand{\thefootnote}{\fnsymbol{footnote}}
    \footnotetext{#1}
    \let\thefootnote=\oldthefootnote
}

\title{Adaptive Contrastive Decoding in Retrieval-Augmented Generation for Handling Noisy Contexts}

\author{
    Youna Kim\textsuperscript{\rm 1},
    Hyuhng Joon Kim\textsuperscript{\rm 1},
    Cheonbok Park\textsuperscript{\rm 2 3},
    Choonghyun Park\textsuperscript{\rm 1},\\
    \textbf{Hyunsoo Cho\textsuperscript{\rm 4},
    Junyeob Kim\textsuperscript{\rm 1},
    Kang Min Yoo\textsuperscript{\rm 1 2 5},
    Sang-goo Lee\textsuperscript{\rm 1 6},
    Taeuk Kim\textsuperscript{\rm 7 *}}\\
    \textsuperscript{\rm 1}Seoul National University,
    \textsuperscript{\rm 2}NAVER Cloud,
    \textsuperscript{\rm 3}KAIST AI,
    \textsuperscript{\rm 4}Ewha Womans University,\\
    \textsuperscript{\rm 5}NAVER AI LAB,
    \textsuperscript{\rm 6}IntelliSys, Korea,
    \textsuperscript{\rm 7}Hanyang University\\
    \{anna9812, heyjoonkim, pch330, juny116, sglee\}@europa.snu.ac.kr\\
    \{cbok.park, kangmin.yoo\}@navercorp.com, chohyunsoo@ewha.ac.kr\\
    kimtaeuk@hanyang.ac.kr\\
}

\begin{document}
\maketitle
    
    \input{Sections/00Abstract}
    \input{Sections/01Introduction}

\input{Sections/02RelatedWork}

    \input{Sections/03Methodology}
    \input{Sections/04ExperimentalSettings}

\input{Sections/051ExperimentalResultsMain}

    \input{Sections/052ExperimentalResultsSub}

    \input{Sections/053ExperimentalResultsAbl1}

\input{Sections/054Knowledgeconflict}

\input{Sections/055Alpha}

    \input{Sections/96Conclusion}
    \input{Sections/97Limitations}

    \input{Sections/98Acknowledgement}

    \bibliography{anthology,custom}
    
    \input{Sections/99Appendix}

\end{document}

%% file: Sections/00Abstract.tex
\begin{abstract}

When using large language models (LLMs) in knowledge-intensive tasks, such as open-domain question answering, external context can bridge the gap between external knowledge and the LLMs' parametric knowledge.
Recent research has been developed to amplify contextual knowledge over the parametric knowledge of LLMs with contrastive decoding approaches.
While these approaches could yield truthful responses when relevant context is provided, they are prone to vulnerabilities when faced with noisy contexts.
We extend the scope of previous studies to encompass noisy contexts and propose adaptive contrastive decoding (\ours) to leverage contextual influence effectively.
\ours\ demonstrates improvements in open-domain question answering tasks compared to baselines, especially in robustness by remaining undistracted by noisy contexts in retrieval-augmented generation.

\astfootnote{Corresponding author.}
\end{abstract}

%% file: Sections/01Introduction.tex
\section{Introduction}

While large language models (LLMs) \citep{touvron2023llama, achiam2023gpt} achieve remarkable performance levels across diverse benchmarks, they sometimes struggle to generalize to knowledge-intensive tasks, such as open-domain question-answering (QA; \citealp{chen-etal-2017-reading}), and may also fail to capture long-tail knowledge, leading to unfaithful output generation \citep{mallen-etal-2023-trust, pmlr-v202-kandpal23a}.
One common approach to address these limitations is fine-tuning the model, but this results in a quadratic rise in computational demands as the size of the LLMs increases exponentially \citep{longpre2023pretrainers}. 
To overcome this, researchers have been investigating strategies to combine non-parametric knowledge with LLMs during response generation without explicit re-training \citep{asai-etal-2023-retrieval}.
This approach leverages external information from knowledge bases and enhances the capability of the LLMs dynamically, ensuring that the information is both current and accurate.

\begin{figure}[t]
\begin{center}
    \includegraphics[width=1\columnwidth]{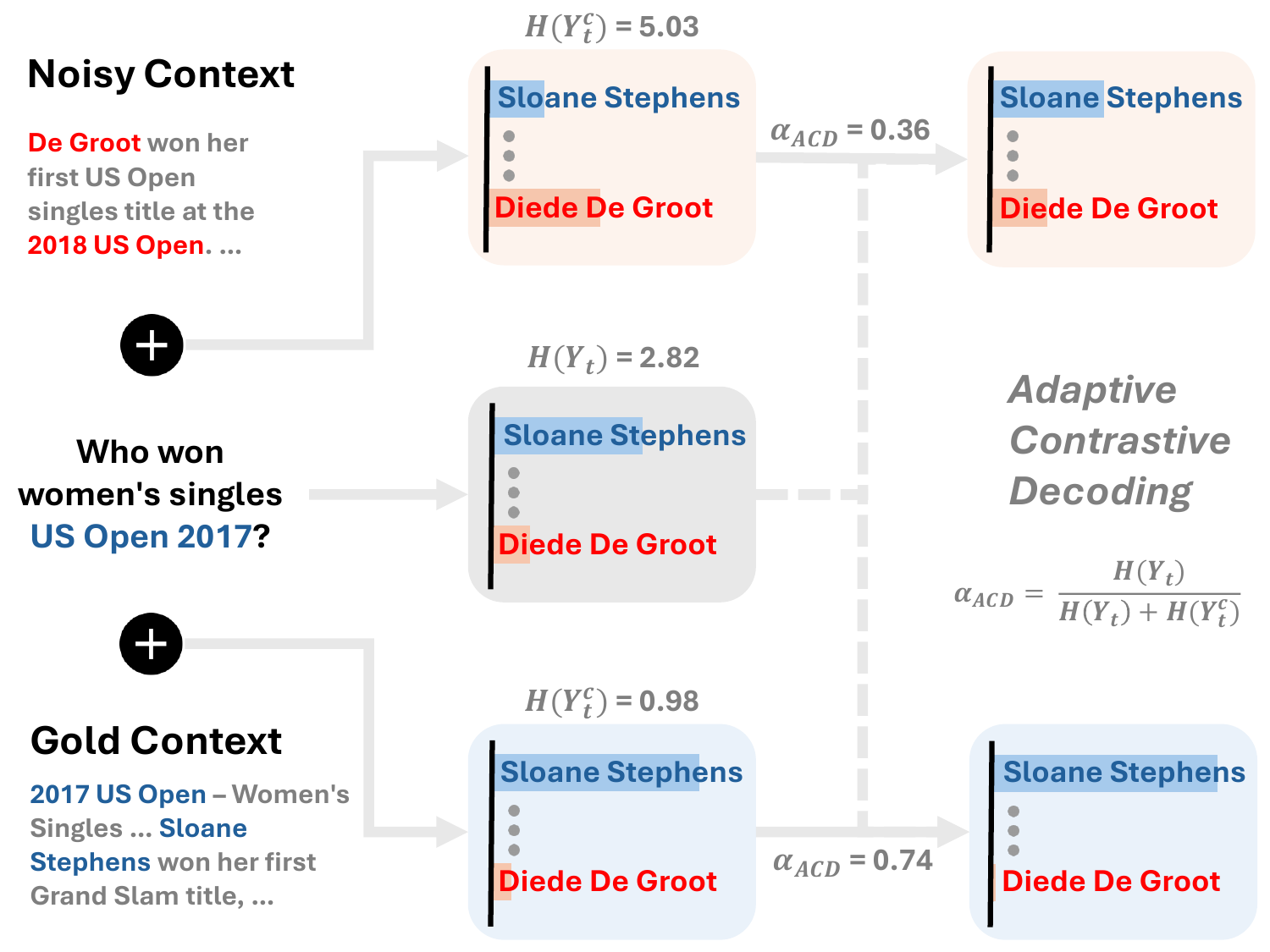}
      \caption{ 
      An illustration of adaptive contrastive decoding (ACD). 
      Entropy ($H$) changes depending on context relevance, affecting the adaptive weight ($\alpha_{\text{ACD}}$).
      Noisy context leads the model to incorrectly answer "Diede De Groot" when employing regular greedy decoding.
      ACD applies context-based adjustments, enabling the correct answer, "Sloane Stephens," despite the noise. 
      }
      \label{figure:front_image}
\end{center}
\end{figure}

Early studies in this field attempt to append query-relevant context to generate more accurate responses.
Especially, contrastive decoding \citep{li-etal-2023-contrastive, malkin-etal-2022-coherence, liu-etal-2021-dexperts} yields significant enhancement in various tasks by amplifying the influence of the given context at decoding step \citep{shi2023trusting, zhao2024enhancing}.  
While such methods work well when context information is correct and faithful, in real-world scenarios, context information is not always correct and may contain some noisy and unfaithful information.
For instance, if the retrieval system pulls in irrelevant or contradictory information, it could lead to incorrect responses \citep{wang2024rear, wu2024easily, yu2024truthaware}. 
This highlights the necessity for a generation model that can gauge the appropriateness of the context by itself, being robust to noise and unfaithful data to ensure the output remains reliable \citep{yoran2024making}.

To assess whether the existing contrastive decoding approaches can be utilized in practice, 
we extend the setting to situations where the gold-standard context is not guaranteed, specifically in the retrieval-augmented generation (RAG) framework \citep{yao2022react, shi2024incontext, JMLR:v24:23-0037}.
In this paper, we demonstrate that existing context-aware contrastive decoding approaches experience performance drops in open-domain question answering, especially when the retrieved context is noisy.
To address this issue, we propose \textbf{adaptive contrastive decoding (\ours)}, adaptively weighting the contrastive contextual influence on the parametric knowledge, making it suitable for noisy context settings (Figure \ref{figure:front_image}).

Incorporating the distinction between contextual and parametric knowledge, our approach aims to mitigate the dominance of potentially noisy contextual information in model output. 
We control contrastive contextual influence based on context's contribution to the LLM's uncertainty reduction, thereby minimizing its disruptive effect during decoding. 
Through in-depth experiments with three open-domain QA datasets, we demonstrate the potential of the proposed approach with increased overall performance.
Moreover, \ours\ enhances the performance significantly on the noisy context scenario while minimizing performance degradation on the gold context scenario compared to the baselines.

%% file: Sections/02RelatedWork.tex
\section{Related Works}

\paragraph{Context-Augmented Generation}
Approaches for context-augmented generation have been developed to enhance the model's limited parametric knowledge by providing external knowledge, enabling more factual and contextually accurate responses during inference \citep{zhou-etal-2023-context, he2024rest}.
To sufficiently incorporate the information from the context in model generation, contrastive decoding approaches are applied to overwrite the model's parametric knowledge with external knowledge \citep{shi2023trusting, zhao2024enhancing}. 
These context-aware contrastive decoding methods to generate responses faithful to the given context show effective performance in summarization \citep{see2017point, narayan-etal-2018-dont}, knowledge conflict \citep{longpre2022entitybased}, and question answering with gold-standard contexts.

\paragraph{Robustness in RAG Frameworks}

While retrieval-augmented generation enables LLMs to become factual and reliable with the retrieved external knowledge, there are still concerns about incorrectly retrieved irrelevant contexts \citep{yoran2024making}.
To address hallucination errors posed by irrelevant contexts, some researchers take an approach to train LLMs that can adaptively retrieve relevant context \citep{asai2023selfrag, wang2024rear}.
Another approach aims to selectively use retrieved contexts after assessing their truthfulness or relevance through context verification with prompting strategies or training untruthful context detectors \citep{yu2024truthaware, zhang2024retrievalqa}. 
These approaches highlight the ongoing efforts to advance the robustness and accuracy of LLMs in multiple directions to manage potentially misleading information.

%% file: Sections/03Methodology.tex
\section{Methodology}

\subsection{Problem Formulation} 
    
    At decoding time step $t$,
    given the input $x$ and preceding sequences $y_{<t}$,
    a pretrained auto-regressive LLM $\theta$ computes the logit $\mathbf{z}_{t} \in \mathbb{R}^{|V|}$, where $V$ is the vocabulary, for the $t$-th token.
    In the open-domain QA task, a question $q$ serves as the input $x$, 
    and $\mathbf{z}_{t}$ relies solely on the LLM's parametric knowledge.
    When both $q$ and the retrieved context $c$ are provided as $x$,
    the logit is denoted as $\mathbf{z}_{t}^c \in \mathbb{R}^{|V|}$.

\subsection{Contrastive Decoding} 

    In cases where context cannot be blindly trusted, directly following the context-augmented distribution  
    can increase the risk of being misled.
    Thus, we adopt the approach of adding the contextual influence, which contrasts with the LLM's parametric knowledge, to the parametric distribution $\mathbf{z}_{t}$.
    With the contrastive decoding objective, $\mathbf{z}_{t}^c$ and $\mathbf{z}_{t}$ are ensembled to reflect the influence of external context on the LLM's parametric knowledge at each decoding step $t$. 
    The probability distribution $P_{\theta}(Y_{t} | x, y_{<t})$ is modified by weighted adjustment based on the difference between $\mathbf{z}_{t}^c$ and $\mathbf{z}_{t}$,
    as represented in the following equation.
    \begin{equation}
        P_{\theta}(Y_{t}\ |\ x, y_{<t}) = \text{softmax}(\mathbf{z}_{t} +\alpha\ (\mathbf{z}_{t}^c - \mathbf{z}_{t}))
    \end{equation}
    The contrastive adjustment enables the LLM to integrate external context $c$ into its prediction, leveraging the weight $\alpha$ to control the impact of $c$ on the final probability distribution.

\subsection{Adaptive Weight on Contextual Influence}

    The degree to which contextual influence is incorporated into $\mathbf{z}_{t}$ needs to be controlled based on the provided context's informativeness.
    In practice, however, it is often unknown whether the context is gold or noisy.
    To address this, we investigate whether the model could adjust accordingly
    with a simple entropy-based approach.

    The LLM's uncertainty is expressed with the entropy $H(Y_{t})$ of its probability distribution $P_{\theta}(Y_{t} | x, y_{<t})$ \citep{huang2023look, kuhn2023semantic}.
    While $ H(Y_{t}) $ reflects how much uncertainty the model has based on its parametric knowledge under the given question,
    $ H(Y_{t}^{c}) $ is influenced by the external knowledge within the retrieved context $c$.
    Generally, when the context is added, the entropy decreases \citep{NIPS2017_2650d608}.
    However, if the context is noisy, irrelevant, or provides no information to answer the given question, it may contribute to increased uncertainty instead.
    
    Intuitively, if the retrieved context provides informative cues for answering the question, then $ H(Y_{t}^{c}) $ is expected to be lowered compared to $ H(Y_{t}) $.
    Conversely, if the context is non-helpful or even confusing the model prediction, $ H(Y_{t}^{c}) $ in predicting the next token is likely to be higher.
    This scenario would be particularly evident when the model knows the answer with low $ H(Y_{t}) $.
    
    Considering the above scenarios, the motivation behind the adaptive weight $\alpha_{\ours}$ is to assign a relatively smaller weight in cases where the context increases uncertainty by being uninformative or confusing for the model in answering the given question.
    Thus, the value of $\alpha_{\ours}$ is set as the proportion of uncertainty contributed by $H(Y_{t})$ relative to the total uncertainty when considering both $H(Y_{t})$ and $H(Y_{t}^{c})$: 
    \begin{equation}
        \alpha_{\ours} = \frac{H(Y_{t})}{H(Y_{t}) + H(Y_{t}^{c})}
    \end{equation}

    Under the condition where $H(Y_{t}) > H(Y_{t}^{c})$, 
    $\alpha_{\ours}$ value approaches to 1,
    indicating that when the context $c$ is provided, the uncertainty associated with predicting the next token decreases. 
    Conversely, when $H(Y_{t}) < H(Y_{t}^{c})$, $\alpha_{\ours}$ value approaches to 0, reflecting minimal influence from $c$.
    Note that when $H(Y_{t}) = H(Y_{t}^{c})$, $\alpha_{\ours}$ becomes 0.5, resulting in an ensemble of two distributions, $\mathbf{z}_{t}$ and $\mathbf{z}_{t}^c$, with equal weighting.
    
    With $\alpha_{\ours}$, the vocab $v$ with maximum probability is selected as the next token under the following distribution: 
    \begin{equation}
        \hat{P_{\theta}}(Y_{t}\ |\ x, y_{<t}) = \text{softmax}(\mathbf{z}_{t}\ +\ \alpha_{\ours}\ (\mathbf{z}_{t}^c - \mathbf{z}_{t}))
    \end{equation}
    
    Informed by $\alpha_{\ours}$ and contextual contrast, 
    the adjustment process determines the degree to which the model's parametric knowledge is superseded, 
    thus optimizing the assimilation of contextual information throughout decoding.

    \input{Assets/Tables/main_results}

%% file: Assets/Tables/main_results.tex
\begin{table*}[tp]
\scriptsize
    \centering
    
\begin{tabular}{ll|ccc|ccc|ccc}
\toprule
 & Dataset ($\rightarrow$) & \multicolumn{3}{c|}{TriviaQA} & \multicolumn{3}{c|}{NQ} & \multicolumn{3}{c}{PopQA}  \\

 \midrule
Model & Method ($\downarrow$) & All & $\text{Subset}_{\text{Gold}}$ & $\text{Subset}_{\text{Noisy}}$ & All & $\text{Subset}_{\text{Gold}}$ & $\text{Subset}_{\text{Noisy}}$ & All & $\text{Subset}_{\text{Gold}}$ & $\text{Subset}_{\text{Noisy}}$ \\

\midrule

\multirow{6}{*}{{\scshape Llama2} 7B} & $\text{Reg}_{Cls}$ & 59.00 & - & - & 25.48 & - & - & 28.36 & - & - \\
 & $\text{Reg}_{Opn}$ & 60.23 & \underline{87.40} & 33.50 & \underline{31.39} & \textbf{61.31} & 12.40 & 38.49 & \underline{81.21} & 7.77 \\
 & CAD & 49.02 & 73.69 & 24.75 & 25.57 & 51.61 & 9.05 & 33.70 & 72.18 & 6.03 \\
 & \cu$_{F}$ & 60.36 & 85.72 & 35.39 & 29.45 & 56.10 & 12.54 & 35.73 & 74.25 & 8.03 \\
 & \cu$_{D}$ & \underline{63.23} & 86.03 & \underline{40.79} & 30.36 & 52.18 & \underline{16.52} & \underline{39.01} & 77.39 & \underline{11.42} \\
 & \ours & \textbf{64.85} & \textbf{88.01} & \textbf{42.06} & \textbf{32.91} & \underline{56.60} & \textbf{17.88} & \textbf{41.29} & \textbf{82.77} & \textbf{11.46} \\

\midrule

\multirow{6}{*}{{\scshape Llama2} 13B} & $\text{Reg}_{Cls}$ & 63.77 & - & - & 30.80 & - & - & 32.70 & - & - \\
 & $\text{Reg}_{Opn}$ & 62.81 & \underline{88.52} & 37.51 & 33.35 & \textbf{62.96} & 14.58 & 40.03 & \underline{83.20} & 8.98 \\
 & CAD & 52.62 & 76.78 & 28.85 & 27.87 & 55.96 & 10.05 & 35.86 & 76.38 & 6.71 \\
 & \cu$_{F}$ & 63.53 & 87.40 & 40.04 & 32.63 & 59.67 & 15.48 & 38.16 & 77.04 & 10.21 \\
 & \cu$_{D}$ & \underline{66.52} & 87.68 & \underline{45.69} & \underline{34.38} & 57.32 & \underline{19.83} & \underline{41.65} & 79.27 & \textbf{14.60} \\
 & \ours & \textbf{67.37} & \textbf{89.36} & \textbf{45.74} & \textbf{36.12} & \underline{61.17} & \textbf{20.24} & \textbf{43.35} & \textbf{83.98} & \underline{14.14} \\

\midrule
\multirow{6}{*}{{\scshape Llama3} 8B} & $\text{Reg}_{Cls}$ & 61.67 & - & - & 28.34 & - & - & 32.65 & - & - \\
& $\text{Reg}_{Opn}$ & 61.27 & \underline{86.94} & 36.02 & \underline{33.30} & \textbf{63.10} & 14.40 & 39.73 & \underline{82.95} & 8.64 \\
& CAD & 49.70 & 72.45 & 27.31 & 29.17 & 58.39 & 10.64 & 35.86 & 76.82 & 6.40 \\
& \cu$_{F}$ & 61.01 & 85.40 & 37.00 & 27.62 & 51.89 & 12.22 & 37.99 & 77.12 & 9.85 \\
& \cu$_{D}$ & \underline{64.01} & 86.08 & \underline{42.28} & 30.72 & 53.96 & \underline{15.98} & \underline{41.35} & 79.32 & \textbf{14.04} \\
& \ours & \textbf{66.32} & \textbf{89.20} & \textbf{43.81} & \textbf{35.48} & \underline{62.03} & \textbf{18.65} & \textbf{43.25} & \textbf{84.48} & \underline{13.60} \\

\midrule
\multirow{6}{*}{{\scshape Mistral} 8B} & $\text{Reg}_{Cls}$ & 63.72 & - & - & 29.64 & - & - & 29.04 & - & - \\
& $\text{Reg}_{Opn}$ & 60.45 & 86.85 & 34.48 & 32.55 & \textbf{64.67} & 12.18 & 38.28 & \underline{81.26} & 7.36 \\
& CAD & 44.69 & 66.89 & 22.85 & 24.10 & 52.25 & 6.25 & 33.93 & 73.95 & 5.15 \\
& \cu$_{F}$ & 63.33 & 88.43 & 38.62 & 31.80 & 61.10 & 13.22 & 36.58 & 76.00 & 8.23 \\
& \cu$_{D}$ & \underline{66.97} & \underline{89.24} & \underline{45.05} & \underline{33.24} & 57.89 & \underline{17.61} & \underline{39.87} & 78.46 & \textbf{12.11} \\
& \ours & \textbf{67.82} & \textbf{90.16} & \textbf{45.83} & \textbf{35.37} & \underline{62.17} & \textbf{18.38} & \textbf{41.47} & \textbf{82.90} & \underline{11.68} \\

\bottomrule
\end{tabular}
\caption{EM accuracy of full data (All) and subsets with gold ($\text{Subset}_{\text{Gold}}$) and noisy contexts ($\text{Subset}_{\text{Noisy}}$). The highest score is in \textbf{bold}, and the second-best is \underline{underlined}.}
\label{table:main_results_main}
\end{table*}

%% file: Sections/04ExperimentalSettings.tex
\section{Experimental Results}

\subsection{Experimental Settings}

\paragraph{Datasets and Models} 
We conduct experiments on open-domain QA datasets, TriviaQA \citep{joshi2017triviaqa}, Natural Questions (NQ; \citealp{kwiatkowski2019natural}), and PopQA \citep{mallen2022not} with Wikipedia contexts.\footnote{Wikipedia dump from Dec. 2018.}

We use auto-regressive language models, {\scshape Llama2} (7B \& 13B, \citealp{touvron2023llama}), {\scshape Llama3 8B},\footnote{\url{https://github.com/meta-llama/llama3}} and {\scshape Mistral 7B} \citep{jiang2023mistral}. Utilizing {\scshape Contriever-msmarco} \citep{izacard2022unsupervised} as a retriever, the top-1 retrieved context is appended to each question.

\paragraph{Evaluation Metric}
Following \citet{zhao2024enhancing}, we use few-shot prompts with 5 examples.
We report Exact Match (EM) as an evaluation metric, which verifies whether the generated sequences precisely match one of the candidate answers.

\paragraph{Baselines}

As fundamental baselines, regular greedy decoding has been employed in open-book ($\text{Reg}_{Opn}$) and closed-book ($\text{Reg}_{Cls}$) settings.
We compare our method against existing context-aware contrastive decoding methods, including Context-Aware Decoding (CAD; \citealp{shi2023trusting}) and Multi-Input Contrastive Decoding (MICD; \citealp{zhao2024enhancing}).
\cu\ uses inputs with and without context, along with an additional input with adversarial context, to generate the output distribution.
\cu\ presents two methods, referred to as \cu$_{F}$ and \cu$_{D}$, which offer fixed and dynamic $\alpha$, respectively.
Similar to our approach, to leverage the burden of hyperparameter search and dependency on fixed $\alpha$, \cu$_{D}$ also determines $\alpha$ dynamically. 
In $\text{\cu}_{D}$, $\alpha$ is assigned as the maximum token probability with context ($\text{max}P_{wc}$) if $\text{max}P_{wc}$ exceeds the maximum token probability without context ($\text{max}P_{woc}$); otherwise, it is calculated as $1-\text{max}P_{woc}$.

%% file: Sections/051ExperimentalResultsMain.tex
\subsection{Main Results}
\label{sec:5_main_results}

\paragraph{Performance on RAG}
As shown in Table \ref{table:main_results_main}, \ours\ outperforms the baselines across all datasets and models within the RAG framework, particularly when considering the full test data (All).
When analyzing the performance by dividing the data into two subsets based on whether the retrieved context is gold ($\text{Subset}_{\text{Gold}}$) or not ($\text{Subset}_{\text{Noisy}}$), \ours\ achieves either the best or second-best performance.
$\text{\cu}_{D}$ demonstrates performance comparable to \ours\ on $\text{Subset}_{\text{Noisy}}$.
However, it shows a significant drop on $\text{Subset}_{\text{Gold}}$, indicating a tendency to ignore gold context while handling noisy context.
It is notable that both CAD and $\text{\cu}_{F}$ exhibit a significant drop in their performance under noisy conditions.

%% file: Sections/052ExperimentalResultsSub.tex
\paragraph{Performance under Parametric Knowledge}
\label{sec:5_ablation1}

\input{Assets/Tables/main_results_figure}

We aim to analyze the model's performance across various aspects, focusing specifically on its parametric knowledge.
We estimate whether the model possesses relevant parametric knowledge for a given question based on its accuracy in a closed-book setting ($\text{Reg}_{Cls}$).
We consider two subsets under the following conditions:
(1) \textit{Known-noisy}: the model has parametric knowledge of the given question and noisy context is retrieved. 
(2) \textit{Unknown-gold}: the model does not have parametric knowledge of the given question and gold context is retrieved.

From Figure \ref{figure:main_results_figure},
we observe that \ours\ outperforms the baselines in \textit{Known-noisy}.
Notably, two approaches with adaptively adjusted weight, \ours\ and $\text{\cu}_{D}$, perform well in \textit{Known-noisy},
while other baselines show a relative strength in \textit{Unknown-gold}.
However, these baselines also experience significant performance drops in \textit{Known-noisy}, indicating distraction by noisy context despite correctly answering when only the question is provided.
In both cases, \ours\ demonstrates better performance compared to $\text{\cu}_{D}$, overall showing a tendency towards reliability.

\input{Assets/Tables/ablation2}

%% file: Assets/Tables/main_results_figure.tex
\begin{figure}[t]
    \centering
    \includegraphics[width=\columnwidth]{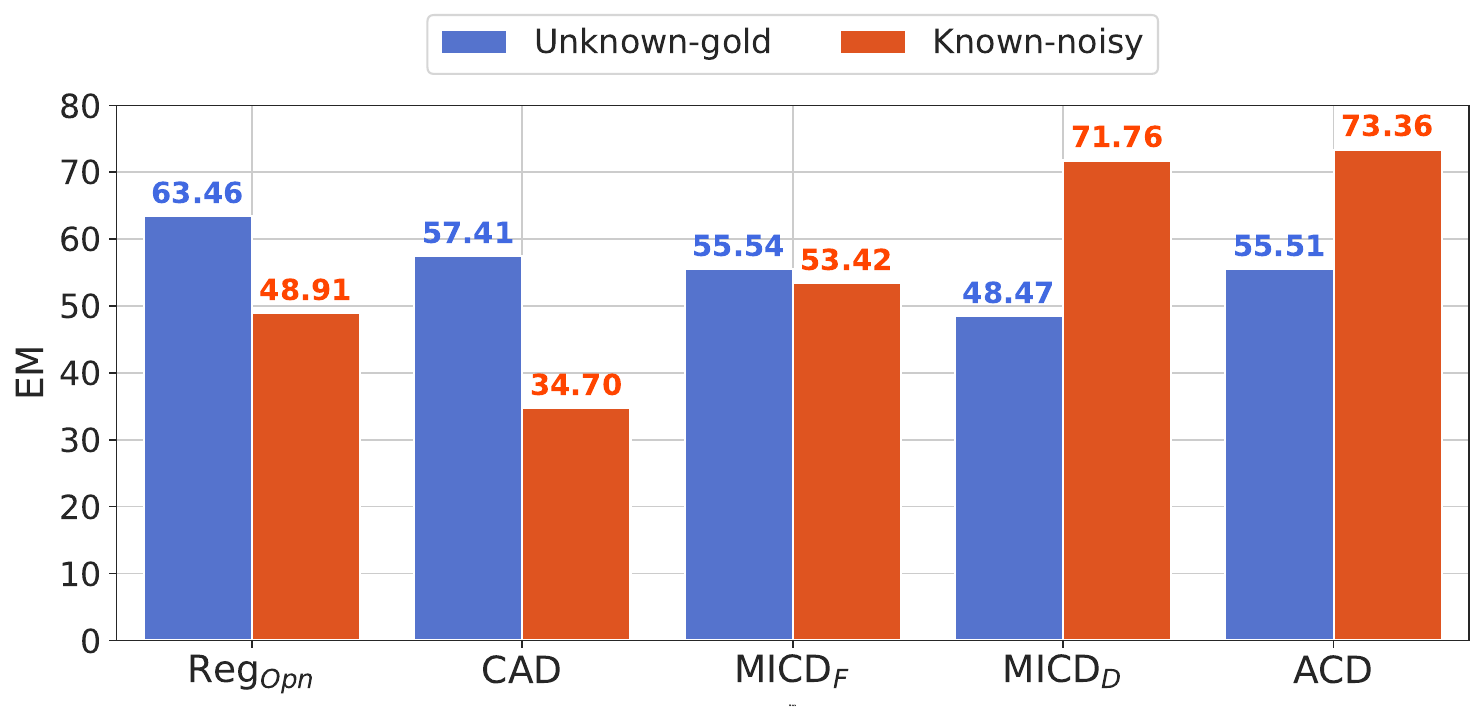}
    \caption{EM accuracy of each method in {\scshape Llama2-7B}. EM of three datasets used are averaged for each subset, \textit{Unknown-gold} and \textit{Known-noisy}.}
    \label{figure:main_results_figure}
\end{figure}

%% file: Assets/Tables/ablation2.tex
\begin{table}[tp]
\centering

\setlength{\tabcolsep}{10pt} 
\resizebox{\columnwidth}{!}{
\begin{tabular}{ll|ccc}
\toprule
 & $\alpha$ & NQ & TriviaQA & PopQA \\
\midrule
\multirow{2}{*}{Max} & \cu$_{\text{D}}$ & 51.53 & 59.76 & 65.49 \\
 & {\cellcolor[rgb]{0.9,0.9,0.9}}\ours & {\cellcolor[rgb]{0.9,0.9,0.9}}\textbf{65.78} & {\cellcolor[rgb]{0.9,0.9,0.9}}\textbf{73.37} & {\cellcolor[rgb]{0.9,0.9,0.9}}\textbf{74.84} \\
\midrule
\multirow{2}{*}{Avg.} & \cu$_{\text{D}}$ & 54.18 & 63.78 & 72.64 \\
 & {\cellcolor[rgb]{0.9,0.9,0.9}}\ours & {\cellcolor[rgb]{0.9,0.9,0.9}}\textbf{68.80} & {\cellcolor[rgb]{0.9,0.9,0.9}}\textbf{72.32} & {\cellcolor[rgb]{0.9,0.9,0.9}}\textbf{78.90} \\
\midrule
\multirow{2}{*}{First} & \cu$_{\text{D}}$ & 53.92 & 62.95 & 68.81 \\
 & {\cellcolor[rgb]{0.9,0.9,0.9}}\ours & {\cellcolor[rgb]{0.9,0.9,0.9}}\textbf{73.27} & {\cellcolor[rgb]{0.9,0.9,0.9}}\textbf{80.45} & {\cellcolor[rgb]{0.9,0.9,0.9}}\textbf{80.08} \\
\bottomrule
\end{tabular}}
\caption{AUROC between $\alpha$ used in each method and the noisiness of the retrieved context.}
\label{table:ablation_alpha}
\end{table}

%% file: Sections/053ExperimentalResultsAbl1.tex
\subsection{Analysis} 

\paragraph{Correlation between Adaptive Weight and Context Noisiness}
    
    While other baselines rely on the fixed hyperparameter of weight $\alpha$, \ours\ and \cu$_{D}$ adjust $\alpha$ during the decoding step. 
    It depends not only on the noisiness of the retrieved context but also on whether the model's parametric knowledge contains an answer to the given question.
    To exclude cases that are not directly related to the analysis of how weight is adjusted based on context quality and the model's parametric knowledge, we use the same subsets, \textit{Known-noisy} and \textit{Unknown-gold}.

    Adaptive weights $\alpha_{\text{\ours}}$ and $\alpha_{\text{\cu}}$ are extracted at each decoding step and analyzed across three metrics: maximum, average, and the first within the generated sequence. 
    As an evaluation metric, the area under the receiver operator characteristic curve (AUROC) between $\alpha$ and the noisiness of the retrieved context is measured.
    AUROC of each $\alpha$ for {\scshape Llama 2-7B} is reported in Table \ref{table:ablation_alpha}.
    Under every metric and dataset, \ours\ demonstrates a higher AUROC compared to \cu$_{\text{D}}$.
    Aligned with our motivation, when the model is knowledgeable and presented with noisy context, $\alpha_{\text{\ours}}$ tends to be lower, emphasizing greater reliance on parametric knowledge.
    Conversely, when the model lacks knowledge and is provided with gold context, $\alpha_{\text{\ours}}$ is adjusted to prioritize reliance on the provided context.

%% file: Sections/054Knowledgeconflict.tex
\paragraph{Handling Knowledge Conflict}
\label{sec:appendix_knowledge_conflict}

\input{Assets/Tables/ablation_nqswap_figure}

With a knowledge conflict QA dataset, NQ-swap \citep{longpre2022entitybased}, we verify whether the two decoding methods with dynamic weight, \ours\ and $\text{\cu}_{D}$, can generate context-based responses without considering a conflicting context as a noisy context.
The conflicting context in the NQ-swap dataset is constructed by replacing the answer entity span in the original gold context with a random entity of the same type.
Figure \ref{figure:abl_nqswap} illustrates that \ours\ consistently exceeds the performance of $\text{\cu}_{D}$ across all models and achieves results comparable to open-book regular decoding.
The results indicate that the \ours's approach remains effective even in settings where the context is relevant to the question but contradicts the model's parametric knowledge.

%% file: Assets/Tables/ablation_nqswap_figure.tex
\begin{figure}[t]
    \centering
    \includegraphics[width=\linewidth]{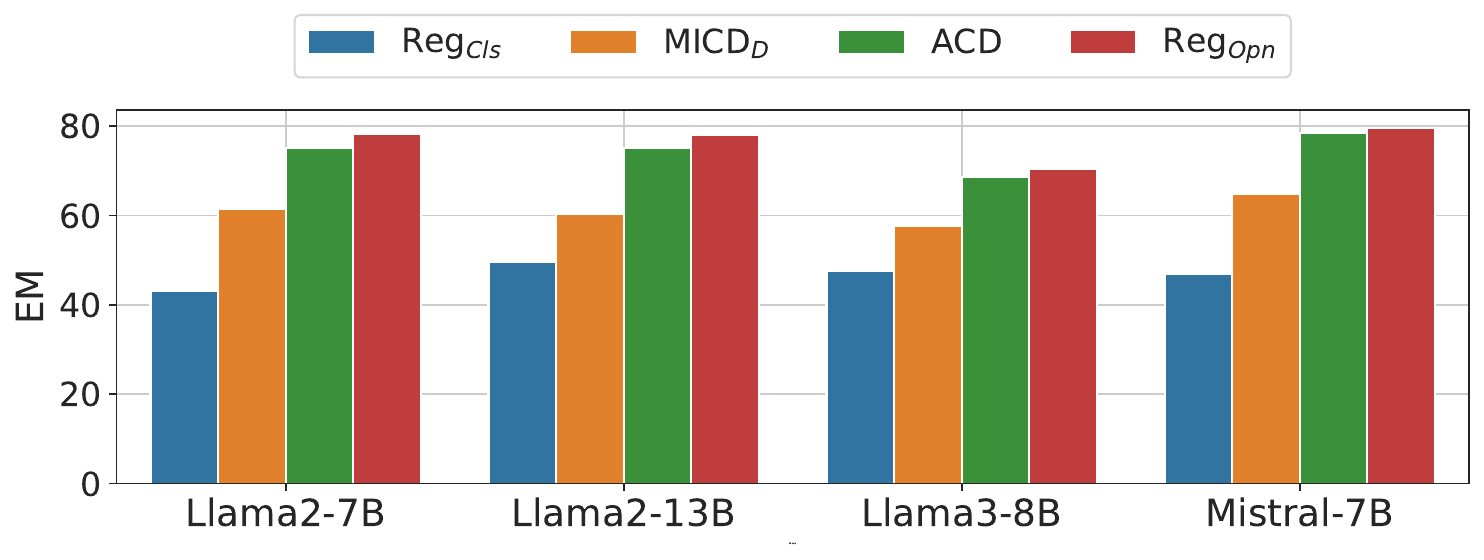}
    \caption{EM accuracy on NQ-swap with contexts replacing the gold answer with a random entity span. }
    \label{figure:abl_nqswap}
\end{figure}

%% file: Sections/055Alpha.tex
\paragraph{Ablation on $\alpha_{\text{\ours}}$}
\input{Assets/Tables/ablation_alpha_figure}

To assess the impact of $\alpha_{\ours}$ on performance, we fix the value of $\alpha$ within a range $[0, 1]$ and examine whether employing \ours\ is more effective than optimizing a fixed weight.
In Figure \ref{figure:abl_alpha}, it can be observed that using a fixed $\alpha$ results in degraded performance compared to \ours. 
Increasing the alpha value, which enhances the contextual influence on the output distribution, initially leads to a rise in the EM score.
However, beyond a certain point, further increasing $\alpha$ results in a decline in the EM score. 
In scenarios with potential noisy context, a fixed $\alpha$ value may not ensure optimal performance.
Therefore, employing an adaptive weight, $\alpha_{\ours}$, to adjust the impact of contextual knowledge based on entropy is crucial for improving overall performance.

%% file: Assets/Tables/ablation_alpha_figure.tex
\begin{figure}[t]
    \centering
    \includegraphics[width=\columnwidth]{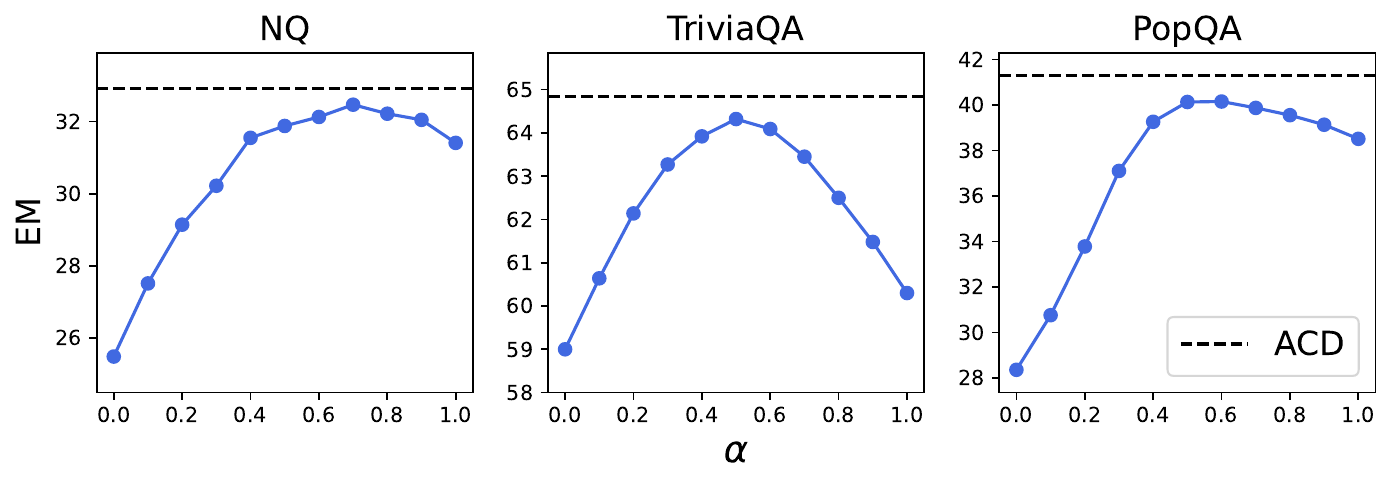}
    \caption{EM across alpha values ranges from 0.0 to 1.0. The dashed line indicates EM score with $\alpha_{ACD}$.}
    \label{figure:abl_alpha}
\end{figure}

%% file: Sections/96Conclusion.tex
\section{Conclusion}

In this work, we mainly tackle handling noisy contexts in open-domain QA on the RAG framework. 
Our proposed method, \ours, dynamically adjusts contextual influence during decoding by quantifying the model's uncertainty that is either reduced or increased by the retrieved context.
Our results show that \ours\ improves performances across various dimensions by considering the LLM's parametric knowledge and context noisiness.
These findings highlight \ours's potential to enhance the reliability of retrieval-augmented generation. 

%% file: Sections/97Limitations.tex
\section*{Limitations}

Similar to other contrastive decoding approaches, the inference cost of our approach is higher than the conventional greedy decoding. 
Specifically, while CAD incurs twice the inference cost and \cu\ incurs three times the cost, \ours\ also incurs twice the inference cost of conventional greedy decoding.

Our research is limited the base models and does not encompass chat or instruction-following models trained with reinforcement learning from human feedback (RLHF) or instruction fine-tuning \citep{ouyang2022training, chung2022scaling}.
These aligned models often generate token distributions that vary significantly based on the presence or absence of contextual instruction or templates.
For instance, an instruction-following model might start its generation with \textit{"According to the given context ..."} when context is provided, while directly generating the answer in absence of context.
This alignment with the provided instructions poses another challenge to be tackled when the contrastive decoding approach is utilized.

Our current focus is primarily on short-form QA tasks.
Expanding to QA tasks with long-form generation will enable a wider range of applications.
Under long-form QA tasks, our approach can be further developed to investigate scenarios where the context is only partially relevant to the question.

%% file: Sections/98Acknowledgement.tex
\section*{Acknowledgement}

This work was partly supported by
SNU-NAVER Hyperscale AI Center and 
Institute of Information \& communications Technology Planning \& Evaluation (IITP) grant funded by the Korea government (MSIT) [NO.RS-2021-II211343, Artificial Intelligence Graduate School Program (Seoul National University), No.RS-2020-II201373, Artificial Intelligence Graduate School Program (Hanyang University), NO.RS-2021-II212068, Artificial Intelligence Innovation Hub (Artificial Intelligence Institute, Seoul National University)]

%% file: Sections/99Appendix.tex
\appendix
\section*{Appendix}

\section{Implementation Details}
\label{sec:appendix_implementaion_details}

\subsection{Instructions}
The templates we use throughout the experiment are in Table \ref{template:without_context} and Table \ref{template:with_context}. The template used in open-book generation (Table \ref{template:with_context}) is applied to get context-augmented distribution $\mathbf{z}_{t}^c$. Also, to obtain $\mathbf{z}_{t}$, the template in Table \ref{template:without_context} is used.

\subsection{Datasets}
For NQ and TriviaQA, general world knowledge is required to answer the given question.
In PopQA, tackling long-tailed information, less popular factual knowledge is asked.
For NQ and TriviaQA, few-shot examples are adopted from train data.
For PopQA, we randomly sample 5 examples with different relationship types for sample diversity.
The number of test data in used is 3,610 for NQ, 11,313 for TriviaQA, and 14,262 for PopQA.

\input{Assets/instruction_wo}

\input{Assets/instruction_w}

\subsection{Baselines}

Baselines using regular greedy decoding are evaluated under two different settings. 
In the closed-book setting, only the question is provided. 
In the open-book setting, the retrieved context is employed.
The same top-1 retrieved context is utilized for every baseline and \ours.

CAD introduces a context-aware contrastive decoding approach that employs a contrastive output distribution to accentuate discrepancies in model predictions with and without context. This method effectively overrides model priors conflicting with provided context, offering significant performance enhancements in tasks requiring resolution of knowledge conflicts.
\cu\ further enhances context grounded generation by integrating contrastive decoding with adversarial irrelevant passages.
From a computational time perspective, \cu\ requires three times more than conventional greedy decoding, while CAD and \ours\ require twice as much.

\cu\ proposes two usage directions, referred to as \cu$_{F}$ and \cu$_{D}$, which offer fixed and dynamic $\alpha$, respectively.
\cu$_{D}$ determines $\alpha$ in use by comparing the highest token probabilities with and without given context.
Throughout the experiments, fixed value of $\alpha$ is set to the value used in \citet{zhao2024enhancing}, 0.5 and 1.0 for CAD and $\text{\cu}_{F}$, respectively.

\subsection{Retriever Performance}
\label{sec:appendix_retriever_performance}

\input{Assets/Tables/retreiver}
To assess performance in the RAG framework, the top-1 context from top-100 contexts retrieved by {\scshape Contriever-msmarco} \citep{izacard2022unsupervised} is utilized.
Recall@100 is reported for each dataset in Table \ref{table:retriever}.

\subsection{Knowledge Conflict}
For the NQ-swap dataset, we utilize the questions and entity-swapped contexts provided in \citet{hong2024hallucinations}, which includes 3,650 samples.
This total excludes 5 few-shot samples and those with contexts presented in a tabular format due to the limited context length.
In the case of NQ-swap, each data point has a given context. 
Since it is a task that does not use a retriever, for \cu, we use the fixed negative context taken from the \cu\ as an adversarial context.
\cu\ reports that the performance difference between fixed negative and the most distant context is negligible.

\section{Results}
\label{sec:appendix_main_results}

\subsection{Results on \textit{Known-noisy} and \textit{Unknown-gold}}
\label{sec:appendix_fr}
For \textit{Known-noisy} and \textit{Unknown-gold}, the exact values of EM accuracy on each case are reported in Table \ref{table:ablation_robustness}  and Table \ref{table:ablation_faithfulness}, respectively.

\subsection{AUROC between Adaptive Weight and Context Noisiness}
\label{sec:appendix_auroc}
AUROC of \ours\ and \cu$_{D}$ for three models not reported in Table \ref{table:ablation_alpha} is reported in Table \ref{table:ablation2_full}.

\section{Additional Analysis}
\label{sec:appendix_additional_analysis}

\subsection{Upper-bound of Alpha}

\input{Assets/Tables/oracle_alpha}

In our approach, the parameter $\alpha$ is expected to be close to 1 when the retrieved context contains information that helps answer the given question, and close to 0 otherwise.
To evaluate the upper-bound performance of \ours, we assume that we have prior knowledge of whether the context in use is gold or noisy. Under this assumption, we fix the $\alpha$ value to 1.0 if the context is gold and to 0.0 if the context is noisy. 

For TriviaQA dataset, the performance of \ours\ is comparable to $\alpha_{oracle}$, with less than 1 point difference (Table \ref{table:oracle_alpha}). NQ and PopQA show a difference of approximately 2-3 points, indicating that the method for calculating the $\alpha$ weight could be further enhanced in future research.

\subsection{Case Study}

\input{Assets/Tables/case_study}

We conduct the case study on $\alpha_{\ours}$, examining its value in cases of \textit{Known-noisy} and \textit{Unknown-gold}.
Table \ref{table:case_study} shows the generations from {\scshape Llama2 7B} and how the values of entropy from closed-book generation ($\text{Reg}_{Cls}$) and open-book generation ($\text{Reg}_{Opn}$) affect $\alpha_{\ours}$ at the first decoding time step.

In the case of \textit{Known-noisy}, when the model generates the answer correctly even without the given context, the retrieved noisy context yields relatively higher entropy, resulting in $\alpha_{\ours}$ value of 0.3483.
Conversely, in the case of \textit{Unknown-gold}, the model's generated answer is incorrect, aligning with a relatively high entropy value of 6.6748.
In this scenario, the retrieved gold context guides the model to correctly answer the question, which is reflected in a relatively lower entropy value of 1.5628.
Thus, the value of $\alpha_{\ours}$, adjusted with these entropy values, yields a relatively higher weight on the context at 0.8103.

\input{Assets/Tables/ablation1-R}
\input{Assets/Tables/ablation1-F}
\input{Assets/Tables/ablation2_full}

%% file: Assets/instruction_wo.tex
\begin{table}[t]
    \centering    

   \begin{tabularx}{\linewidth}{X}
    \toprule
    
\ttfamily
Answer the following questions: \\\\
\ttfamily
\textcolor{brown}{<few-shots>} \\\\
\ttfamily
Question: \textcolor{brown}{<question>} \\
\ttfamily
Answer:
\\
    \bottomrule
\end{tabularx}

    \caption{Template used in closed-book generation.}
    \label{template:without_context}
\end{table}

%% file: Assets/instruction_w.tex
\begin{table}[t]
    \centering    

   \begin{tabularx}{\linewidth}{X}
    \toprule
    
\ttfamily
Answer the following questions: \\\\
\ttfamily
\textcolor{brown}{<few-shots>} \\\\
\ttfamily
Context: \textcolor{brown}{<context>} \\
\ttfamily
Question: \textcolor{brown}{<question>} \\
\ttfamily
Answer:
\\
    \bottomrule
\end{tabularx}

    \caption{Template used in open-book generation.}
    \label{template:with_context}
\end{table}

%% file: Assets/Tables/retreiver.tex
\begin{table}[tp]
\resizebox{\columnwidth}{!}{
\begin{tabular}{l|ccccc}
\toprule
 & R@1 & R@5 & R@10 & R@20 & R@100 \\
\midrule
NQ & 38.81 & 65.65 & 73.91 & 79.56 & 88.01 \\
TriviaQA & 49.60 & 71.32 & 76.72 & 80.39 & 85.71 \\
PopQA & 41.83 & 61.54 & 68.63 & 74.55 & 83.95 \\
\bottomrule
\end{tabular}}
\caption{Recall@100 performance for {\scshape Contriever-msmarco}}
\label{table:retriever}
\end{table}

%% file: Assets/Tables/oracle_alpha.tex
\definecolor{best}{HTML}{5469FF}
\definecolor{least}{HTML}{FF5454}

\begin{table}[tp]
\centering
\setlength{\tabcolsep}{10pt}
\resizebox{\columnwidth}{!}{
\begin{tabular}{lccc}
\toprule
 & \multicolumn{1}{c}{NQ} & \multicolumn{1}{c}{TriviaQA} & \multicolumn{1}{c}{PopQA} \\

\midrule
\multicolumn{4}{c}{{\scshape Llama2-7B}} \\
\midrule
$\alpha_{\ours}$ & 32.91 & 64.85 & 41.29 \\
$\alpha_{oracle}$ & 35.35 \textcolor{best}{(+2.44)} & 65.31 \textcolor{best}{(+0.46)} & 44.10 \textcolor{best}{(+2.81)}\\

\midrule
\multicolumn{4}{c}{{\scshape Llama2-13B}} \\
\midrule
$\alpha_{\ours}$ & 36.12 & 67.37 & 43.35 \\
$\alpha_{oracle}$ & 38.75 \textcolor{best}{(+2.63)} & 68.19 \textcolor{best}{(+0.82)} & 47.01 \textcolor{best}{(+3.66)} \\

\midrule
\multicolumn{4}{c}{{\scshape Llama3 8B}} \\
\midrule
$\alpha_{\ours}$ & 35.48 & 66.32 & 43.25 \\
$\alpha_{oracle}$ & 36.98 \textcolor{best}{(+1.50)} & 66.10 \textcolor{least}{(-0.22)} & 46.47 \textcolor{best}{(+3.22)} \\

\midrule
\multicolumn{4}{c}{{\scshape Mistral 7B}} \\
\midrule
$\alpha_{\ours}$ & 35.37 & 67.82 & 41.47 \\
$\alpha_{oracle}$ & 38.37 \textcolor{best}{(+3.00)}& 67.29  \textcolor{least}{(-0.53)} & 44.53 \textcolor{best}{(+3.06)} \\

\bottomrule

\end{tabular}}

\caption{EM score comparison between \ours\ ($\alpha_{\ours}$) and \ours\ with oracle alpha value ($\alpha_{oracle}$).}
\label{table:oracle_alpha}
\end{table}

%% file: Assets/Tables/case_study.tex
\begin{table*}[tp]
\scriptsize
\centering    
\begin{tabular}{ll|cc|cc|cc}
\toprule
 & Sample & \multicolumn{2}{c|}{$\text{Reg}_{Cls}$} & \multicolumn{2}{c|}{$\text{Reg}_{Opn}$} & \multicolumn{2}{c}{\ours} \\
 \midrule
Case &  & \multicolumn{1}{c}{Generation} & \multicolumn{1}{c|}{$H(Y_{t})$} & \multicolumn{1}{c}{Generation} & \multicolumn{1}{c|}{$H(Y_{t}^{c})$} & \multicolumn{1}{c}{Generation} & \multicolumn{1}{l}{$\alpha_{\ours}$}  \\
\midrule
\textit{Known-noisy} & Question: who does the voice & \multirow{3}{*}{Moira Kelly} & \multirow{3}{*}{2.9160} & \multirow{3}{*}{Whoopi Goldberg} & \multirow{3}{*}{5.4562} & \multirow{3}{*}{Moira Kelly} & \multirow{3}{*}{0.3483} \\
 & of nala in the lion king? &  &  &  &  &  &  \\
 & Gold answer: Moira Kelly &  &  &  &  &  & \\

 \midrule
\textit{Unknown-gold} & Question: who was the actor that & \multirow{3}{*}{Michael Tucker} & \multirow{3}{*}{6.6748} & \multirow{3}{*}{Michael Moriarty} & \multirow{3}{*}{1.5628} & \multirow{3}{*}{Michael Moriarty} & \multirow{3}{*}{0.8103}  \\
 & played ben stone on law and order? &  &  &  &  &  &  \\
 & Gold answer: Michael Moriarty &  &  &  &  &  & \\
\bottomrule

\end{tabular}
\caption{Case study on the value of $\alpha_{\ours}$ for \textit{Known-noisy} and \textit{Unknown-gold} cases in {\scshape Llama2 7B}. Each value of entropy without context ($H(Y_{t})$), entropy with context ($H(Y_{t}^{c})$), and $\alpha_{\ours}$ is extracted at the first decoding step ($t=0$).}
\label{table:case_study}
\end{table*}

%% file: Assets/Tables/ablation1-R.tex
\begin{table}[tp]
\centering
\begin{tabular}{l|ccc}
\toprule
 & NQ & TriviaQA & PopQA \\
\midrule
\multicolumn{4}{c}{{\scshape Llama2-7B}} \\
\midrule
$\text{Reg}_{Opn}$ & 45.13 & 68.12 & 33.47 \\
CAD & 29.22 & 48.91 & 25.97 \\
\cu$_{F}$ & 51.07 & 72.37 & 36.81 \\
\cu$_{D}$ & \underline{72.92} & \underline{86.33} & \textbf{56.04} \\
\ours & \textbf{76.72} & \textbf{88.79} & \underline{54.58} \\
\midrule
\multicolumn{4}{c}{{\scshape Llama2-13B}} \\
\midrule
$\text{Reg}_{Opn}$ & 47.18 & 69.77 & 32.53 \\
CAD & 32.04 & 52.48 & 22.66 \\
\cu$_{F}$ & 54.17 & 75.05 & 38.55 \\
\cu$_{D}$ & \textbf{76.31} & \underline{88.24} & \textbf{59.38} \\
\ours & \underline{75.15} & \textbf{88.78} & \underline{56.11} \\
\midrule
\multicolumn{4}{c}{{\scshape Llama3-8B}} \\
\midrule
$\text{Reg}_{Opn}$ & 46.20 & 68.50 & 33.39 \\
CAD & 32.91 & 50.51 & 23.00 \\
\cu$_{F}$ & 43.25 & 70.67 & 39.07 \\
\cu$_{D}$ & \underline{61.18} & \underline{83.70} & \textbf{59.80} \\
\ours & \textbf{64.14} & \textbf{86.59} & \underline{56.87} \\
\midrule
\multicolumn{4}{c}{{\scshape Mistral-7B}} \\
\midrule
$\text{Reg}_{Opn}$ & 41.04 & 64.57 & 31.03 \\
CAD & 19.17 & 42.63 & 20.80 \\
\cu$_{F}$ & 48.12 & 71.99 & 36.48 \\
\cu$_{D}$ & \underline{69.58} & \underline{86.84} & \textbf{57.14} \\
\ours & \textbf{70.62} & \textbf{89.36} & \underline{53.55} \\
\bottomrule
\end{tabular}
\caption{EM accuracy of \textit{Known-noisy} case. }
\label{table:ablation_robustness}
\end{table}

%% file: Assets/Tables/ablation1-F.tex
\begin{table}[tp]
\centering
\begin{tabular}{l|ccc}
\toprule
 & NQ & TriviaQA & PopQA \\
\midrule
\multicolumn{4}{c}{{\scshape Llama2-7B}} \\
\midrule
$\text{Reg}_{Opn}$ & \textbf{47.78} & \textbf{68.18} & \textbf{74.42} \\
CAD & \underline{43.90} & \underline{62.22} & 66.12 \\
\cu$_{F}$ & 40.47 & 61.51 & 64.63 \\
\cu$_{D}$ & 29.82 & 50.43 & 65.17 \\
\ours & 36.03 & 57.10 & \underline{73.41} \\
\midrule
\multicolumn{4}{c}{{\scshape Llama2-13B}} \\
\midrule
$\text{Reg}_{Opn}$ & \textbf{46.52} & \textbf{65.09} & \textbf{75.04} \\
CAD & \underline{45.77} & 61.07 & 69.43 \\
\cu$_{F}$ & 41.79 & \underline{62.03} & 65.85 \\
\cu$_{D}$ & 30.72 & 47.77 & 64.70 \\
\ours & 36.19 & 53.98 & \underline{72.38} \\
\midrule
\multicolumn{4}{c}{{\scshape Llama3-8B}} \\
\midrule
$\text{Reg}_{Opn}$ & \textbf{48.12} & \textbf{68.47} & \textbf{74.10} \\
CAD & \underline{48.00} & 60.52 & 70.41 \\
\cu$_{F}$ & 38.15 & \underline{61.40} & 65.55 \\
\cu$_{D}$ & 33.33 & 48.45 & 64.81 \\
\ours & 41.67 & 61.24 & \underline{72.52} \\
\midrule
\multicolumn{4}{c}{{\scshape Mistral-7B}} \\
\midrule
$\text{Reg}_{Opn}$ & \textbf{49.57} & \textbf{64.82} & \textbf{73.09} \\
CAD & \underline{45.38} & 56.02 & 67.81 \\
\cu$_{F}$ & 43.28 & \underline{63.59} & 66.58 \\
\cu$_{D}$ & 32.06 & 54.97 & 66.37 \\
\ours & 37.73 & 57.70 & \underline{73.03} \\
\bottomrule
\end{tabular}
\caption{EM accuracy of \textit{Unknown-gold} case. }
\label{table:ablation_faithfulness}
\end{table}

%% file: Assets/Tables/ablation2_full.tex
\begin{table}[tp]
\centering
\resizebox{\columnwidth}{!}{
\begin{tabular}{l|l|ccc}
\toprule
 & \textbf{$\alpha$} & NQ & TriviaQA & PopQA \\
 
 \midrule
\multicolumn{5}{c}{{\scshape Llama2 13B}} \\
 \midrule
\multirow{2}{*}{Max} & \cu$_{\text{D}}$ & 52.77 & 60.09 & 61.84 \\
 & \ours & \textbf{69.24} & \textbf{75.31} & \textbf{74.12} \\
\midrule
\multirow{2}{*}{Avg.} & \cu$_{\text{D}}$ & 57.86 & 62.00 & 71.79 \\
 & \ours & \textbf{71.61} & \textbf{73.41} & \textbf{77.92} \\
\midrule
\multirow{2}{*}{First} & \cu$_{\text{D}}$ & 54.80 & 46.13 & 68.44 \\
 & \ours & \textbf{73.07} & \textbf{77.96} & \textbf{80.51} \\
 \midrule
\multicolumn{5}{c}{{\scshape Llama3 8B}} \\
\midrule
\multirow{2}{*}{Max} & \cu$_{\text{D}}$ & 50.75 & 52.59 & 63.72 \\
 & \ours & \textbf{63.12} & \textbf{57.82} & \textbf{75.00} \\
\midrule
\multirow{2}{*}{Avg.} & \cu$_{\text{D}}$ & 51.80 & 52.83 & 67.99 \\
 & \ours & \textbf{64.08} & \textbf{59.67} & \textbf{75.90} \\
\midrule
\multirow{2}{*}{First} & \cu$_{\text{D}}$ & 45.70 & 39.07 & 69.21 \\
 & \ours & \textbf{67.48} & \textbf{75.45} & \textbf{80.31} \\
 \midrule
\multicolumn{5}{c}{{\scshape Mistral 7B}} \\
\midrule
\multirow{2}{*}{Max} & \cu$_{\text{D}}$ & 56.98 & 64.95 & 61.93 \\
 & \ours & \textbf{71.27} & \textbf{77.46} & \textbf{74.11} \\
\midrule
\multirow{2}{*}{Avg.} & \cu$_{\text{D}}$ & 63.66 & 69.27 & 73.82 \\
 & \ours & \textbf{76.02} & \textbf{78.20} & \textbf{79.08} \\
\midrule
\multirow{2}{*}{First} & \cu$_{\text{D}}$ & 56.84 & 68.98 & 71.73 \\
 & \ours & \textbf{75.75} & \textbf{84.11} & \textbf{82.07} \\
\bottomrule
\end{tabular}}
\caption{AUROC between $\alpha$ used in each method and the noisiness of the retrieved context. The best AUROC is in bold. }
\label{table:ablation2_full}
\end{table}

%% file: main.bbl
\begin{thebibliography}{37}
\providecommand{\natexlab}[1]{#1}

\bibitem[{Achiam et~al.(2023)Achiam, Adler, Agarwal, Ahmad, Akkaya, Aleman, Almeida, Altenschmidt, Altman, Anadkat et~al.}]{achiam2023gpt}
Josh Achiam, Steven Adler, Sandhini Agarwal, Lama Ahmad, Ilge Akkaya, Florencia~Leoni Aleman, Diogo Almeida, Janko Altenschmidt, Sam Altman, Shyamal Anadkat, et~al. 2023.
\newblock Gpt-4 technical report.
\newblock \emph{arXiv preprint arXiv:2303.08774}.

\bibitem[{Asai et~al.(2023{\natexlab{a}})Asai, Min, Zhong, and Chen}]{asai-etal-2023-retrieval}
Akari Asai, Sewon Min, Zexuan Zhong, and Danqi Chen. 2023{\natexlab{a}}.
\newblock \href {https://doi.org/10.18653/v1/2023.acl-tutorials.6} {Retrieval-based language models and applications}.
\newblock In \emph{Proceedings of the 61st Annual Meeting of the Association for Computational Linguistics (Volume 6: Tutorial Abstracts)}, pages 41--46, Toronto, Canada. Association for Computational Linguistics.

\bibitem[{Asai et~al.(2023{\natexlab{b}})Asai, Wu, Wang, Sil, and Hajishirzi}]{asai2023selfrag}
Akari Asai, Zeqiu Wu, Yizhong Wang, Avirup Sil, and Hannaneh Hajishirzi. 2023{\natexlab{b}}.
\newblock \href {https://arxiv.org/abs/2310.11511} {Self-rag: Learning to retrieve, generate, and critique through self-reflection}.
\newblock \emph{Preprint}, arXiv:2310.11511.

\bibitem[{Chen et~al.(2017)Chen, Fisch, Weston, and Bordes}]{chen-etal-2017-reading}
Danqi Chen, Adam Fisch, Jason Weston, and Antoine Bordes. 2017.
\newblock \href {https://doi.org/10.18653/v1/P17-1171} {Reading {W}ikipedia to answer open-domain questions}.
\newblock In \emph{Proceedings of the 55th Annual Meeting of the Association for Computational Linguistics (Volume 1: Long Papers)}, pages 1870--1879, Vancouver, Canada. Association for Computational Linguistics.

\bibitem[{Chung et~al.(2022)Chung, Hou, Longpre, Zoph, Tay, Fedus, Li, Wang, Dehghani, Brahma, Webson, Gu, Dai, Suzgun, Chen, Chowdhery, Castro-Ros, Pellat, Robinson, Valter, Narang, Mishra, Yu, Zhao, Huang, Dai, Yu, Petrov, Chi, Dean, Devlin, Roberts, Zhou, Le, and Wei}]{chung2022scaling}
Hyung~Won Chung, Le~Hou, Shayne Longpre, Barret Zoph, Yi~Tay, William Fedus, Yunxuan Li, Xuezhi Wang, Mostafa Dehghani, Siddhartha Brahma, Albert Webson, Shixiang~Shane Gu, Zhuyun Dai, Mirac Suzgun, Xinyun Chen, Aakanksha Chowdhery, Alex Castro-Ros, Marie Pellat, Kevin Robinson, Dasha Valter, Sharan Narang, Gaurav Mishra, Adams Yu, Vincent Zhao, Yanping Huang, Andrew Dai, Hongkun Yu, Slav Petrov, Ed~H. Chi, Jeff Dean, Jacob Devlin, Adam Roberts, Denny Zhou, Quoc~V. Le, and Jason Wei. 2022.
\newblock \href {https://arxiv.org/abs/2210.11416} {Scaling instruction-finetuned language models}.
\newblock \emph{Preprint}, arXiv:2210.11416.

\bibitem[{He et~al.(2024)He, Zhong, Cai, Lee, and He}]{he2024rest}
Zhenyu He, Zexuan Zhong, Tianle Cai, Jason~D. Lee, and Di~He. 2024.
\newblock \href {https://arxiv.org/abs/2311.08252} {Rest: Retrieval-based speculative decoding}.
\newblock \emph{Preprint}, arXiv:2311.08252.

\bibitem[{Hong et~al.(2024)Hong, Gema, Saxena, Du, Nie, Zhao, Perez-Beltrachini, Ryabinin, He, Fourrier, and Minervini}]{hong2024hallucinations}
Giwon Hong, Aryo~Pradipta Gema, Rohit Saxena, Xiaotang Du, Ping Nie, Yu~Zhao, Laura Perez-Beltrachini, Max Ryabinin, Xuanli He, Clémentine Fourrier, and Pasquale Minervini. 2024.
\newblock \href {https://arxiv.org/abs/2404.05904} {The hallucinations leaderboard -- an open effort to measure hallucinations in large language models}.
\newblock \emph{Preprint}, arXiv:2404.05904.

\bibitem[{Huang et~al.(2023)Huang, Song, Wang, Zhao, Chen, Juefei-Xu, and Ma}]{huang2023look}
Yuheng Huang, Jiayang Song, Zhijie Wang, Shengming Zhao, Huaming Chen, Felix Juefei-Xu, and Lei Ma. 2023.
\newblock \href {https://arxiv.org/abs/2307.10236} {Look before you leap: An exploratory study of uncertainty measurement for large language models}.
\newblock \emph{Preprint}, arXiv:2307.10236.

\bibitem[{Izacard et~al.(2022)Izacard, Caron, Hosseini, Riedel, Bojanowski, Joulin, and Grave}]{izacard2022unsupervised}
Gautier Izacard, Mathilde Caron, Lucas Hosseini, Sebastian Riedel, Piotr Bojanowski, Armand Joulin, and Edouard Grave. 2022.
\newblock \href {https://arxiv.org/abs/2112.09118} {Unsupervised dense information retrieval with contrastive learning}.
\newblock \emph{Preprint}, arXiv:2112.09118.

\bibitem[{Izacard et~al.(2023)Izacard, Lewis, Lomeli, Hosseini, Petroni, Schick, Dwivedi-Yu, Joulin, Riedel, and Grave}]{JMLR:v24:23-0037}
Gautier Izacard, Patrick Lewis, Maria Lomeli, Lucas Hosseini, Fabio Petroni, Timo Schick, Jane Dwivedi-Yu, Armand Joulin, Sebastian Riedel, and Edouard Grave. 2023.
\newblock \href {http://jmlr.org/papers/v24/23-0037.html} {Atlas: Few-shot learning with retrieval augmented language models}.
\newblock \emph{Journal of Machine Learning Research}, 24(251):1--43.

\bibitem[{Jiang et~al.(2023)Jiang, Sablayrolles, Mensch, Bamford, Chaplot, de~las Casas, Bressand, Lengyel, Lample, Saulnier, Lavaud, Lachaux, Stock, Scao, Lavril, Wang, Lacroix, and Sayed}]{jiang2023mistral}
Albert~Q. Jiang, Alexandre Sablayrolles, Arthur Mensch, Chris Bamford, Devendra~Singh Chaplot, Diego de~las Casas, Florian Bressand, Gianna Lengyel, Guillaume Lample, Lucile Saulnier, Lélio~Renard Lavaud, Marie-Anne Lachaux, Pierre Stock, Teven~Le Scao, Thibaut Lavril, Thomas Wang, Timothée Lacroix, and William~El Sayed. 2023.
\newblock \href {https://arxiv.org/abs/2310.06825} {Mistral 7b}.
\newblock \emph{Preprint}, arXiv:2310.06825.

\bibitem[{Joshi et~al.(2017)Joshi, Choi, Weld, and Zettlemoyer}]{joshi2017triviaqa}
Mandar Joshi, Eunsol Choi, Daniel~S. Weld, and Luke Zettlemoyer. 2017.
\newblock \href {https://arxiv.org/abs/1705.03551} {Triviaqa: A large scale distantly supervised challenge dataset for reading comprehension}.
\newblock \emph{Preprint}, arXiv:1705.03551.

\bibitem[{Kandpal et~al.(2023)Kandpal, Deng, Roberts, Wallace, and Raffel}]{pmlr-v202-kandpal23a}
Nikhil Kandpal, Haikang Deng, Adam Roberts, Eric Wallace, and Colin Raffel. 2023.
\newblock \href {https://proceedings.mlr.press/v202/kandpal23a.html} {Large language models struggle to learn long-tail knowledge}.
\newblock In \emph{Proceedings of the 40th International Conference on Machine Learning}, volume 202 of \emph{Proceedings of Machine Learning Research}, pages 15696--15707. PMLR.

\bibitem[{Kendall and Gal(2017)}]{NIPS2017_2650d608}
Alex Kendall and Yarin Gal. 2017.
\newblock \href {https://proceedings.neurips.cc/paper_files/paper/2017/file/2650d6089a6d640c5e85b2b88265dc2b-Paper.pdf} {What uncertainties do we need in bayesian deep learning for computer vision?}
\newblock In \emph{Advances in Neural Information Processing Systems}, volume~30. Curran Associates, Inc.

\bibitem[{Kuhn et~al.(2023)Kuhn, Gal, and Farquhar}]{kuhn2023semantic}
Lorenz Kuhn, Yarin Gal, and Sebastian Farquhar. 2023.
\newblock \href {https://arxiv.org/abs/2302.09664} {Semantic uncertainty: Linguistic invariances for uncertainty estimation in natural language generation}.
\newblock \emph{Preprint}, arXiv:2302.09664.

\bibitem[{Kwiatkowski et~al.(2019)Kwiatkowski, Palomaki, Redfield, Collins, Parikh, Alberti, Epstein, Polosukhin, Devlin, Lee et~al.}]{kwiatkowski2019natural}
Tom Kwiatkowski, Jennimaria Palomaki, Olivia Redfield, Michael Collins, Ankur Parikh, Chris Alberti, Danielle Epstein, Illia Polosukhin, Jacob Devlin, Kenton Lee, et~al. 2019.
\newblock Natural questions: a benchmark for question answering research.
\newblock \emph{Transactions of the Association for Computational Linguistics}, 7:453--466.

\bibitem[{Li et~al.(2023)Li, Holtzman, Fried, Liang, Eisner, Hashimoto, Zettlemoyer, and Lewis}]{li-etal-2023-contrastive}
Xiang~Lisa Li, Ari Holtzman, Daniel Fried, Percy Liang, Jason Eisner, Tatsunori Hashimoto, Luke Zettlemoyer, and Mike Lewis. 2023.
\newblock \href {https://doi.org/10.18653/v1/2023.acl-long.687} {Contrastive decoding: Open-ended text generation as optimization}.
\newblock In \emph{Proceedings of the 61st Annual Meeting of the Association for Computational Linguistics (Volume 1: Long Papers)}, pages 12286--12312, Toronto, Canada. Association for Computational Linguistics.

\bibitem[{Liu et~al.(2021)Liu, Sap, Lu, Swayamdipta, Bhagavatula, Smith, and Choi}]{liu-etal-2021-dexperts}
Alisa Liu, Maarten Sap, Ximing Lu, Swabha Swayamdipta, Chandra Bhagavatula, Noah~A. Smith, and Yejin Choi. 2021.
\newblock \href {https://doi.org/10.18653/v1/2021.acl-long.522} {{DE}xperts: Decoding-time controlled text generation with experts and anti-experts}.
\newblock In \emph{Proceedings of the 59th Annual Meeting of the Association for Computational Linguistics and the 11th International Joint Conference on Natural Language Processing (Volume 1: Long Papers)}, pages 6691--6706, Online. Association for Computational Linguistics.

\bibitem[{Longpre et~al.(2022)Longpre, Perisetla, Chen, Ramesh, DuBois, and Singh}]{longpre2022entitybased}
Shayne Longpre, Kartik Perisetla, Anthony Chen, Nikhil Ramesh, Chris DuBois, and Sameer Singh. 2022.
\newblock \href {https://arxiv.org/abs/2109.05052} {Entity-based knowledge conflicts in question answering}.
\newblock \emph{Preprint}, arXiv:2109.05052.

\bibitem[{Longpre et~al.(2023)Longpre, Yauney, Reif, Lee, Roberts, Zoph, Zhou, Wei, Robinson, Mimno, and Ippolito}]{longpre2023pretrainers}
Shayne Longpre, Gregory Yauney, Emily Reif, Katherine Lee, Adam Roberts, Barret Zoph, Denny Zhou, Jason Wei, Kevin Robinson, David Mimno, and Daphne Ippolito. 2023.
\newblock \href {https://arxiv.org/abs/2305.13169} {A pretrainer's guide to training data: Measuring the effects of data age, domain coverage, quality, \& toxicity}.
\newblock \emph{Preprint}, arXiv:2305.13169.

\bibitem[{Malkin et~al.(2022)Malkin, Wang, and Jojic}]{malkin-etal-2022-coherence}
Nikolay Malkin, Zhen Wang, and Nebojsa Jojic. 2022.
\newblock \href {https://doi.org/10.18653/v1/2022.acl-long.565} {Coherence boosting: When your pretrained language model is not paying enough attention}.
\newblock In \emph{Proceedings of the 60th Annual Meeting of the Association for Computational Linguistics (Volume 1: Long Papers)}, pages 8214--8236, Dublin, Ireland. Association for Computational Linguistics.

\bibitem[{Mallen et~al.(2022)Mallen, Asai, Zhong, Das, Khashabi, and Hajishirzi}]{mallen2022not}
Alex Mallen, Akari Asai, Victor Zhong, Rajarshi Das, Daniel Khashabi, and Hannaneh Hajishirzi. 2022.
\newblock When not to trust language models: Investigating effectiveness of parametric and non-parametric memories.
\newblock \emph{arXiv preprint arXiv:2212.10511}.

\bibitem[{Mallen et~al.(2023)Mallen, Asai, Zhong, Das, Khashabi, and Hajishirzi}]{mallen-etal-2023-trust}
Alex Mallen, Akari Asai, Victor Zhong, Rajarshi Das, Daniel Khashabi, and Hannaneh Hajishirzi. 2023.
\newblock \href {https://doi.org/10.18653/v1/2023.acl-long.546} {When not to trust language models: Investigating effectiveness of parametric and non-parametric memories}.
\newblock In \emph{Proceedings of the 61st Annual Meeting of the Association for Computational Linguistics (Volume 1: Long Papers)}, pages 9802--9822, Toronto, Canada. Association for Computational Linguistics.

\bibitem[{Narayan et~al.(2018)Narayan, Cohen, and Lapata}]{narayan-etal-2018-dont}
Shashi Narayan, Shay~B. Cohen, and Mirella Lapata. 2018.
\newblock \href {https://doi.org/10.18653/v1/D18-1206} {Don{'}t give me the details, just the summary! topic-aware convolutional neural networks for extreme summarization}.
\newblock In \emph{Proceedings of the 2018 Conference on Empirical Methods in Natural Language Processing}, pages 1797--1807, Brussels, Belgium. Association for Computational Linguistics.

\bibitem[{Ouyang et~al.(2022)Ouyang, Wu, Jiang, Almeida, Wainwright, Mishkin, Zhang, Agarwal, Slama, Ray, Schulman, Hilton, Kelton, Miller, Simens, Askell, Welinder, Christiano, Leike, and Lowe}]{ouyang2022training}
Long Ouyang, Jeff Wu, Xu~Jiang, Diogo Almeida, Carroll~L. Wainwright, Pamela Mishkin, Chong Zhang, Sandhini Agarwal, Katarina Slama, Alex Ray, John Schulman, Jacob Hilton, Fraser Kelton, Luke Miller, Maddie Simens, Amanda Askell, Peter Welinder, Paul Christiano, Jan Leike, and Ryan Lowe. 2022.
\newblock \href {https://arxiv.org/abs/2203.02155} {Training language models to follow instructions with human feedback}.
\newblock \emph{Preprint}, arXiv:2203.02155.

\bibitem[{See et~al.(2017)See, Liu, and Manning}]{see2017point}
Abigail See, Peter~J. Liu, and Christopher~D. Manning. 2017.
\newblock \href {https://arxiv.org/abs/1704.04368} {Get to the point: Summarization with pointer-generator networks}.
\newblock \emph{Preprint}, arXiv:1704.04368.

\bibitem[{Shi et~al.(2023)Shi, Han, Lewis, Tsvetkov, Zettlemoyer, and tau Yih}]{shi2023trusting}
Weijia Shi, Xiaochuang Han, Mike Lewis, Yulia Tsvetkov, Luke Zettlemoyer, and Scott~Wen tau Yih. 2023.
\newblock \href {https://arxiv.org/abs/2305.14739} {Trusting your evidence: Hallucinate less with context-aware decoding}.
\newblock \emph{Preprint}, arXiv:2305.14739.

\bibitem[{Shi et~al.(2024)Shi, Min, Lomeli, Zhou, Li, Szilvasy, James, Lin, Smith, Zettlemoyer, Yih, and Lewis}]{shi2024incontext}
Weijia Shi, Sewon Min, Maria Lomeli, Chunting Zhou, Margaret Li, Gergely Szilvasy, Rich James, Xi~Victoria Lin, Noah~A. Smith, Luke Zettlemoyer, Scott Yih, and Mike Lewis. 2024.
\newblock \href {https://arxiv.org/abs/2310.10638} {In-context pretraining: Language modeling beyond document boundaries}.
\newblock \emph{Preprint}, arXiv:2310.10638.

\bibitem[{Touvron et~al.(2023)Touvron, Martin, Stone, Albert, Almahairi, Babaei, Bashlykov, Batra, Bhargava, Bhosale, Bikel, Blecher, Ferrer, Chen, Cucurull, Esiobu, Fernandes, Fu, Fu, Fuller, Gao, Goswami, Goyal, Hartshorn, Hosseini, Hou, Inan, Kardas, Kerkez, Khabsa, Kloumann, Korenev, Koura, Lachaux, Lavril, Lee, Liskovich, Lu, Mao, Martinet, Mihaylov, Mishra, Molybog, Nie, Poulton, Reizenstein, Rungta, Saladi, Schelten, Silva, Smith, Subramanian, Tan, Tang, Taylor, Williams, Kuan, Xu, Yan, Zarov, Zhang, Fan, Kambadur, Narang, Rodriguez, Stojnic, Edunov, and Scialom}]{touvron2023llama}
Hugo Touvron, Louis Martin, Kevin Stone, Peter Albert, Amjad Almahairi, Yasmine Babaei, Nikolay Bashlykov, Soumya Batra, Prajjwal Bhargava, Shruti Bhosale, Dan Bikel, Lukas Blecher, Cristian~Canton Ferrer, Moya Chen, Guillem Cucurull, David Esiobu, Jude Fernandes, Jeremy Fu, Wenyin Fu, Brian Fuller, Cynthia Gao, Vedanuj Goswami, Naman Goyal, Anthony Hartshorn, Saghar Hosseini, Rui Hou, Hakan Inan, Marcin Kardas, Viktor Kerkez, Madian Khabsa, Isabel Kloumann, Artem Korenev, Punit~Singh Koura, Marie-Anne Lachaux, Thibaut Lavril, Jenya Lee, Diana Liskovich, Yinghai Lu, Yuning Mao, Xavier Martinet, Todor Mihaylov, Pushkar Mishra, Igor Molybog, Yixin Nie, Andrew Poulton, Jeremy Reizenstein, Rashi Rungta, Kalyan Saladi, Alan Schelten, Ruan Silva, Eric~Michael Smith, Ranjan Subramanian, Xiaoqing~Ellen Tan, Binh Tang, Ross Taylor, Adina Williams, Jian~Xiang Kuan, Puxin Xu, Zheng Yan, Iliyan Zarov, Yuchen Zhang, Angela Fan, Melanie Kambadur, Sharan Narang, Aurelien Rodriguez, Robert Stojnic, Sergey Edunov, and Thomas
  Scialom. 2023.
\newblock \href {https://arxiv.org/abs/2307.09288} {Llama 2: Open foundation and fine-tuned chat models}.
\newblock \emph{Preprint}, arXiv:2307.09288.

\bibitem[{Wang et~al.(2024)Wang, Ren, Li, Zhao, Liu, and Wen}]{wang2024rear}
Yuhao Wang, Ruiyang Ren, Junyi Li, Wayne~Xin Zhao, Jing Liu, and Ji-Rong Wen. 2024.
\newblock \href {https://arxiv.org/abs/2402.17497} {Rear: A relevance-aware retrieval-augmented framework for open-domain question answering}.
\newblock \emph{Preprint}, arXiv:2402.17497.

\bibitem[{Wu et~al.(2024)Wu, Xie, Chen, Zhu, Zhang, and Xiao}]{wu2024easily}
Siye Wu, Jian Xie, Jiangjie Chen, Tinghui Zhu, Kai Zhang, and Yanghua Xiao. 2024.
\newblock \href {https://arxiv.org/abs/2404.03302} {How easily do irrelevant inputs skew the responses of large language models?}
\newblock \emph{Preprint}, arXiv:2404.03302.

\bibitem[{Yao et~al.(2022)Yao, Zhao, Yu, Du, Shafran, Narasimhan, and Cao}]{yao2022react}
Shunyu Yao, Jeffrey Zhao, Dian Yu, Nan Du, Izhak Shafran, Karthik Narasimhan, and Yuan Cao. 2022.
\newblock React: Synergizing reasoning and acting in language models.
\newblock \emph{arXiv preprint arXiv:2210.03629}.

\bibitem[{Yoran et~al.(2024)Yoran, Wolfson, Ram, and Berant}]{yoran2024making}
Ori Yoran, Tomer Wolfson, Ori Ram, and Jonathan Berant. 2024.
\newblock \href {https://arxiv.org/abs/2310.01558} {Making retrieval-augmented language models robust to irrelevant context}.
\newblock \emph{Preprint}, arXiv:2310.01558.

\bibitem[{Yu et~al.(2024)Yu, Zhang, and Feng}]{yu2024truthaware}
Tian Yu, Shaolei Zhang, and Yang Feng. 2024.
\newblock \href {https://arxiv.org/abs/2403.07556} {Truth-aware context selection: Mitigating hallucinations of large language models being misled by untruthful contexts}.
\newblock \emph{Preprint}, arXiv:2403.07556.

\bibitem[{Zhang et~al.(2024)Zhang, Fang, and Chen}]{zhang2024retrievalqa}
Zihan Zhang, Meng Fang, and Ling Chen. 2024.
\newblock \href {https://arxiv.org/abs/2402.16457} {Retrievalqa: Assessing adaptive retrieval-augmented generation for short-form open-domain question answering}.
\newblock \emph{Preprint}, arXiv:2402.16457.

\bibitem[{Zhao et~al.(2024)Zhao, Monti, Lehmann, and Assem}]{zhao2024enhancing}
Zheng Zhao, Emilio Monti, Jens Lehmann, and Haytham Assem. 2024.
\newblock \href {https://arxiv.org/abs/2405.02750} {Enhancing contextual understanding in large language models through contrastive decoding}.
\newblock \emph{Preprint}, arXiv:2405.02750.

\bibitem[{Zhou et~al.(2023)Zhou, Zhang, Poon, and Chen}]{zhou-etal-2023-context}
Wenxuan Zhou, Sheng Zhang, Hoifung Poon, and Muhao Chen. 2023.
\newblock \href {https://doi.org/10.18653/v1/2023.findings-emnlp.968} {Context-faithful prompting for large language models}.
\newblock In \emph{Findings of the Association for Computational Linguistics: EMNLP 2023}, pages 14544--14556, Singapore. Association for Computational Linguistics.

\end{thebibliography}
